\documentclass[letterpaper, 10 pt, conference]{ieeeconf}  

\IEEEoverridecommandlockouts                              

\usepackage{amsmath,amsfonts,amssymb}
\usepackage{graphicx}
\usepackage{makecell}
\usepackage{mathtools}
\usepackage{subcaption}
\usepackage[table]{xcolor}

\makeatletter
\newcommand{\vast}{\bBigg@{4}}
\newcommand{\Vast}{\bBigg@{5}}
\makeatother

\DeclareCaptionLabelSeparator{periodspace}{.\quad}
\title{\LARGE \bf
Learning Interactive Behaviors for Musculoskeletal Robots Using Bayesian Interaction Primitives
}

\author{Joseph Campbell$^{1}$, Arne Hitzmann$^{2}$, Simon Stepputtis$^{1}$, Shuhei Ikemoto$^{3}$, Koh Hosoda$^{2}$, and Heni Ben Amor$^{1}$
\thanks{$^{1}$J.~Campbell, S.~Stepputtis, and H.~Ben~Amor are with the school of Computing, Informatics, and Decision Systems Engineering, Arizona State University
        {\tt\small \{jacampb1, sstepput, hbenamor\}@asu.edu}}%
\thanks{$^{2}$A.~Hitzmann and K.~Hosoda are with the Graduate School of Engineering Science, Osaka University
        {\tt\small \{arne.hitzmann, hosoda\}@sys.es.osaka-u.ac.jp}}%
\thanks{$^{3}$S.~Ikemoto is with the Graduate School of Life Science and Systems Engineering, Kyushu
Institute of Technology
        {\tt\small ikemoto@brain.kyutech.ac.jp}}%
}

\begin{document}

\maketitle
\thispagestyle{empty}
\pagestyle{empty}

\begin{abstract}
Musculoskeletal robots that are based on pneumatic actuation have a variety of properties, such as compliance and back-drivability, that render them particularly appealing for human-robot collaboration. However, programming interactive and responsive behaviors for such systems is extremely challenging due to the nonlinearity and uncertainty inherent to their control. In this paper, we propose an approach for learning Bayesian Interaction Primitives for musculoskeletal robots given a limited set of example demonstrations.
We show that this approach is capable of real-time state estimation and response generation for interaction with a robot for which no analytical model exists.
Human-robot interaction experiments on a 'handshake' task show that the approach generalizes to new positions, interaction partners, and movement velocities.
\end{abstract}

\section{Introduction}
As robots are employed in an ever-expanding variety of roles, interactions with humans will inevitably become common place. To ensure that such interactions are safe and productive, robots need to be both mechanically and behaviorally responsive to their human counterparts. Mechanical responsiveness and compliance guarantees that no physical contact or force-exchange is harmful to the human. Traditional robotic systems are typically composed of non-compliant, rigid limbs that do not yield when contact with an opposing body occurs. In contrast to that, various musculoskeletal robots have been proposed which are based on McKibben pneumatic actuators~\cite{klute1999mckibben}. These biologically-inspired systems mimic the behavior of human muscles and tendons, and are capable of providing a significant amount of force while remaining inherently safe due to their compliant, back-drivable nature. Unfortunately, due to the uncertainty and nonlinearity underlying pneumatic actuation, controlling such systems can pose major difficulties to the control framework~\cite{tondu2005seven,van2009proxy}. Often, getting a complex robot to perform smooth, generalizable actions on its own can be extremely challenging, let alone react to a human interaction partner. 
\begin{figure}[ht!]
  \begin{center}
    \includegraphics[width=0.85\linewidth]{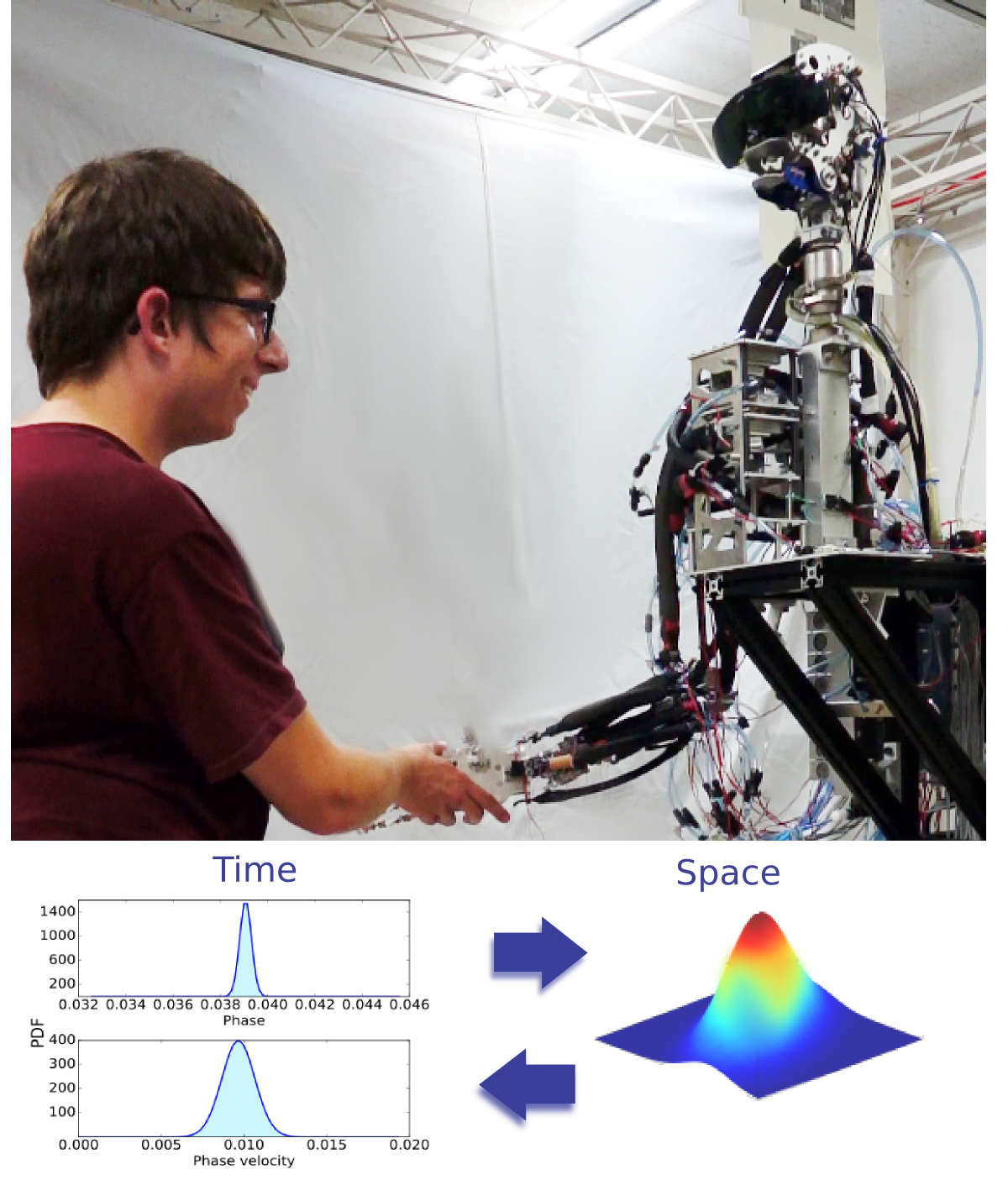}
  \end{center}
  \caption{A musculoskeletal robot learning to shake the hand of a human partner. Bayesian Interaction Primitives are used to determine \emph{when} and \emph{how} to interact.\label{fig:teaser}}
  \label{fig:teaser}
\end{figure}

In this paper, we leverage a methodology for teaching interactive and responsive behaviors to musculoskeletal robots known as Bayesian Interaction Primitives (BIP)~\cite{campbell2017a}.
Based on \emph{learning from demonstration}, this framework provides the basis for working with robots that have no tractable analytical model.
Given a small set of training demonstrations, a spatiotemporal model is extracted which correlates the movements of the interaction partners.
This information is efficiently encoded in a joint probability distribution which, in turn, can be used to infer \emph{when} and \emph{how} to engage in collaborative actions with a human interaction partner.

However, despite the promise of such a framework, Bayesian Interaction Primitives have yet to be successfully demonstrated in a real-time human-robot interaction scenario, and it has yet to be shown whether this framework is capable of generalizing to new interaction partners.
Furthermore, while BIP introduced a new integrated temporal estimation technique which differed from prior work~\cite{amor2014interaction, maeda2014learning}, there is no detailed analysis of its effectiveness nor operating characteristics.
This work intends to fill that gap with the following contributions.
Specifically, we:
\begin{itemize}
    \item introduce a methodology for learning from demonstration in human-robot interaction scenarios which combines open-loop robot trajectories with reactive human behavior in order to generate a responsive robot control policy,
    \item demonstrate that our approach can effectively generate legible, temporally- and spatially-adaptive response trajectories for a complex, musculoskeletal robot for which no analytical model exists,
    \item empirically analyze the integrated phase estimation of our approach, including how well it adapts to different interaction speeds, starting times, and edge cases, as well as how this impacts uncertainty estimates,
    \item demonstrate its ability to generalize to new interaction partners unseen in the training process.
\end{itemize}
In the next section we will discuss the existing literature, and the gap that exists in learning algorithms for human-robot interaction with pneumatic robots that we aim to address.
Section~\ref{sec:methodology} provides an overview of our approach while Sec.~\ref{sec:results} describes our experimental setup and provides an extensive analysis of the results.
Section~\ref{sec:conclusion} outlines our findings and the directions for our future work.

\section{Related Work}

Robots with pneumatic artificial muscles (PAMs) and compliant limbs have been shown to be desirable for human-robot interaction scenarios~\cite{noritsugu1997application,tsagarakis2003development}.
When configured in an anthropomorphic musculoskeletal structure, such robots provide an intriguing platform for human-robot interaction (HRI)~\cite{thomaz2016computational} due to their potential to generate human-like motions while offering a degree of safety as a result of their compliance when confronted with an external force -- such as contact with a human.
Recent work~\cite{hitzmann2018anthropomorphic} has shown the value of utilizing McKibben actuators in the design of these robots, due to their inherent compliance and inexpensive material cost.
However, while analytical kinematics models are in theory possible~\cite{chou1994static, ikemoto2012direct}, they are not always practical due to the effects of friction and the deterioration of mechanical elements, which are difficult to account for (although some gains have been made in this area~\cite{hildebrandt2005cascaded}).
Subsequently, this work proposes using a method based on learning from demonstration~\cite{billard2008robot, argall2009survey}, which is a well-established methodology for teaching robots complex motor skills based on observed data.

A particularly prominent technique in this regard is the Dynamic Movement Primitives~\cite{schaal2006dynamic} framework in which sensor trajectories are approximated by a nonlinear dynamical system. After training, the parameters of the dynamical system can be changed so as to generalize the observed skill to novel scenarios. The extension to human-robot interaction via Interaction Primitives~\cite{amor2014interaction} opened the door to complex interactions with humans that are capable of generalizing to a wide variety of scenarios. A probabilistic framing of Interaction Primitives~\cite{maeda2014learning} replaces the approximation with a probability distribution over a weighted combination of linear basis functions, further strengthening the generalization capabilities despite requiring an additional controller to ensure safe execution of trajectories. The representation involving probability distributions allows for training mixture models~\cite{ewerton2015learning}, thereby enabling the effective learning of multiple complex actions. However, phase estimation and weight inference are performed separately, thus failing to leverage all of the prior information. Recent theoretical work on Interaction Primitives has addressed this limitation by proposing a fully Bayesian derivation of the underlying concepts~\cite{campbell2017a}.
Yet these algorithms have largely only been deployed in traditional robots with electromechanical actuators and rigid limbs~\cite{kulic2012incremental,rozo2016learning}.
This work aims to fill that gap by showing that sufficiently robust learning algorithms can safely and accurately enable human-robot interaction in the face of nonlinear dynamics imposed by musculoskeletal systems.

\section{Bayesian Interaction Primitives}
\label{sec:methodology}

Bayesian Interaction Primitives (BIP)~\cite{campbell2017a} are a novel HRI framework based on learning from demonstration. Intuitively, BIP models the actions of two agents -- a human and a robot -- as time dependent trajectories for each measured degree of freedom.
Demonstration trajectories are represented as a weighted superposition of linear basis models. In turn, the relationship between actions can be captured by computing the covariance between basis weights.
During run-time, an observed trajectory is generated from one of the agents which is localized in both time and space.
Once localized, the trajectory for the other agent can be generated based on the relationship learned from the demonstrations.

\subsection{Interaction Latent Model}
\label{sec:prelim_basis_decomp}

We define an interaction $\boldsymbol{Y}$ as a time series of $D$-dimensional sensor observations over time, $\boldsymbol{Y}_{1:T} = [\boldsymbol{y}_1, \dots, \boldsymbol{y}_T] \in \mathbb{R}^{D \times T}$. Of the $D$ dimensions, $D_o$ of them represent \emph{observed} DoFs from the human and $D_c$ of them represent the \emph{controlled} DoFs from the robot, such that $D = D_c + D_o$.

In order to decouple the size of the state space from the number of observations while maintaining the shape of the trajectories, we transform the interaction $\boldsymbol{Y}$ into a latent space via basis function decomposition.
Each dimension $d \in D$ of $\boldsymbol{Y}$ is approximated with a weighted linear combination of time-dependent basis functions: $[y^d_{1}, \dots, y^d_{t}] = [\Phi_{\phi(1)}^{d} \boldsymbol{w}^d + \epsilon_y, \dots, \Phi_{\phi(t)}^{d} \boldsymbol{w}^d + \epsilon_y]$, where $\Phi_{\phi(t)}^{d} \in \mathbb{R}^{1 \times B^d}$ is a row vector of $B^d$ basis functions, $\boldsymbol{w}^d \in \mathbb{R}^{B^d \times 1}$, and $\epsilon_y$ is i.i.d. Gaussian noise.
In this work, we employ Gaussian basis functions as they are widely used in this type of application~\cite{maeda2014learning}.
Given that this forms a linear system of equations, linear regression is employed to find the weights $\boldsymbol{w}^d$.
The weights from each dimension are aggregated together to form the full latent model of the interaction, $\boldsymbol{w} = [\boldsymbol{w}^{1\intercal}, \dots, \boldsymbol{w}^{D\intercal}] \in \mathbb{R}^{1 \times B}$ where $B = \sum_{d}^{D} B^d$.

The above basis functions are dependent on a relative phase value, $\phi(t)$, rather than the absolute time $t$, such that the range of the phase function is linearly interpolated from $[0, 1]$ over the domain $[0, T]$.
The purpose of phase is to decouple the shape of a trajectory from its speed; when transformed into phase space, a trajectory performed at both a slow and fast movement speed will yield the same basis function decomposition.
In subsequent equations, we will refer to $\phi(t)$ as simply $\phi$ to reduce notational clutter.

\subsection{Spatiotemporal Filtering}
\label{sec:prelim_filtering}

The objective of BIP is to infer the latent model of an interaction, $\boldsymbol{w}$, given a prior model $\boldsymbol{w}_0$ and a partial observation $\boldsymbol{Y}_{1:t}$ where $\phi(t) < 1$.
We assume $T$ is unknown, and so we must also simultaneously estimate the phase of the interaction at the same time as the latent model.
This is possible due to correlated errors in the weights of the latent model stemming from a shared error in the phase estimate~\cite{campbell2017a}.
Intuitively, an error in the temporal estimate will produce an error in the spatial estimate.
We model this simultaneous inference of space and time by augmenting the state vector with both the phase and the phase velocity -- the speed of the interaction -- such that $\boldsymbol{s} = [\phi, \dot{\phi}, \boldsymbol{w}]$ and
\begin{equation}
\label{eq:bip_general}
p(\boldsymbol{s}_t | \boldsymbol{Y}_{1:t}, \boldsymbol{s}_{0}) \propto p(\boldsymbol{y}_{t} | \boldsymbol{s}_t) p(\boldsymbol{s}_t | \boldsymbol{Y}_{1:t-1}, \boldsymbol{s}_{0}).
\end{equation}
It is important to note that while the weights themselves are time-invariant with respect to an interaction, our estimate of the weights \emph{is} time-varying.

The posterior density in Eq.~\ref{eq:bip_general} is computed with a recursive linear state space filter~\cite{thrun2005probabilistic} which consists of two steps: the propagation of the state forward in time according to the system dynamics $p(\boldsymbol{s}_t | \boldsymbol{Y}_{1:t-1}, \boldsymbol{s}_{0})$, and the update of the state based on the latest sensor observation likelihood $p(\boldsymbol{y}_{t} | \boldsymbol{s}_t)$.
We assume this system satisfies the Markov property, such that the state prediction density is defined as:
\begin{align}
\label{eq:bip_prediction_reduced}
& p(\boldsymbol{s}_t | \boldsymbol{Y}_{1:t-1}, \boldsymbol{s}_{0}) \nonumber \\
& = \int p(\boldsymbol{s}_t | \boldsymbol{s}_{t-1}) %
p(\boldsymbol{s}_{t-1} | \boldsymbol{Y}_{1:t-1}, \boldsymbol{s}_{0})d\boldsymbol{s}_{t-1}.
\end{align}
Furthermore, as in the Kalman filter, we assume that the uncertainty associated with our state estimate is normally distributed, i.e., $p(\boldsymbol{s}_t | \boldsymbol{Y}_{1:t}, \boldsymbol{s}_{0}) = \mathcal{N}(\boldsymbol{\mu}_{t|t}, \boldsymbol{\Sigma}_{t|t})$ and $p(\boldsymbol{s}_t | \boldsymbol{Y}_{1:t-1}, \boldsymbol{s}_{0}) = \mathcal{N}(\boldsymbol{\mu}_{t|t-1}, \boldsymbol{\Sigma}_{t|t-1})$.
The system dynamics are defined such that the state evolves according to a linear constant velocity model:
\begin{align}
\boldsymbol{\mu}_{t|t-1} &= 
{\underbrace{
		\begin{bmatrix}
		1 & \Delta t & \dots & 0\\
		0 & 1 & \dots & 0\\
		\vdots & \vdots & \ddots & \vdots\\
		0 & 0 & \dots & 1
		\end{bmatrix}
	}_\text{$\boldsymbol{G}$}}
\boldsymbol{\mu}_{t-1|t-1},\label{eq:ip_state_transition}\\
\boldsymbol{\Sigma}_{t|t-1} &= \boldsymbol{G} \boldsymbol{\Sigma}_{t-1|t-1} \boldsymbol{G}^{\intercal} +
\underbrace{\begin{bmatrix}
	\Sigma_{\phi, \phi} & \Sigma_{\phi, \dot{\phi}} & \dots & 0\\
	\Sigma_{\dot{\phi}, \phi} & \Sigma_{\dot{\phi}, \dot{\phi}} & \dots & 0\\
	\vdots & \vdots & \ddots & \vdots\\
	0 & 0 & \dots & 1
	\end{bmatrix}}_\text{$\boldsymbol{Q}_t$},
\label{eq:ip_cov_initial}
\end{align}
where $\boldsymbol{Q}_t$ is the process noise associated with the state transition, e.g., discrete white noise.
The observation function $h(\cdot)$ is nonlinear and linearized via Taylor expansion:
\begin{align}
\begin{split}
\boldsymbol{H}_t &= \frac{\partial h(\boldsymbol{s}_t)}{\partial s_t}
= 
\begin{bmatrix}
\frac{\partial \Phi_{\phi}^{1} \boldsymbol{w}^1}{\partial \phi} & 0 & \Phi_{\phi}^{1} & \dots & 0\\
\vdots & \vdots & \vdots & \ddots & \vdots\\
\frac{\partial \Phi_{\phi}^{D} \boldsymbol{w}^{D}}{\partial \phi} & 0 & 0 & \dots & \Phi_{\phi}^{D}
\end{bmatrix}. \label{eq:bip_jacobian}
\end{split}
\end{align}
This yields the measurement update equations
\begin{align}
\boldsymbol{K}_t &= \boldsymbol{\Sigma}_{t|t-1} \boldsymbol{H}_t^{\intercal} (\boldsymbol{H}_t \boldsymbol{\Sigma}_{t|t-1} \boldsymbol{H}_t^{\intercal} + \boldsymbol{R}_t)^{-1},\\
\boldsymbol{\mu}_{t|t} &= \boldsymbol{\mu}_{t|t-1} + \boldsymbol{K}_t(\boldsymbol{y}_t - h(\boldsymbol{\mu}_{t|t-1})),\\
\boldsymbol{\Sigma}_{t|t} &= (I - \boldsymbol{K}_t \boldsymbol{H}_t)\boldsymbol{\Sigma}_{t|t-1}, \label{eq:bip_cov_update}
\end{align}
where $\boldsymbol{R}_t$ is the Gaussian measurement noise associated with the sensor observation $\boldsymbol{y}_t$.

We compute the prior model $\boldsymbol{s}_0 = [\phi_0, \dot{\phi}_0, \boldsymbol{w}_0]$ from a set of $N$ initial demonstrations, $\boldsymbol{W} = [\boldsymbol{w}_1^{\intercal}, \dots, \boldsymbol{w}_N^{\intercal}]$, such that $\boldsymbol{w}_0$ is the arithmetic mean of the weights from each DoF:
\begin{equation}
\label{eq:ip_prior}
\boldsymbol{w}_0 = \left[\frac{1}{N}\sum_{i=1}^{N}\boldsymbol{w}^1_i, \dots, \frac{1}{N}\sum_{i=1}^{N}\boldsymbol{w}^D_i\right].
\end{equation}
Assuming that all interactions start at the beginning, we set the initial phase $\phi_0$ to $0$. 
The initial phase velocity $\dot{\phi}_0$, however, is determined by the arithmetic mean of the set of phase velocities from all demonstrations:
\begin{equation}
\dot{\phi}_0 = \frac{1}{N} \sum_{i=1}^N \frac{1}{T_i},
\label{eq:phase_vel_prior}
\end{equation}
where $T_i$ is the length of the $i$-th demonstration.
Lastly, we define the prior density $p(\boldsymbol{s}_0) = \mathcal{N}(\boldsymbol{\mu}_0, \boldsymbol{\Sigma}_0)$ as
\begin{align}
\boldsymbol{\mu}_0 &= \boldsymbol{s}_0, \qquad
\boldsymbol{\Sigma}_0 = %
\begin{bmatrix}
\boldsymbol{\Sigma}_{\phi, \phi} & 0\\
0 & \boldsymbol{\Sigma}_{\boldsymbol{W}, \boldsymbol{W}}
\end{bmatrix}, \label{eq:bip_prior_cov}
\end{align}
where $\boldsymbol{\Sigma}_{\phi, \phi}$ is the variance in the phases and phase velocities of the demonstrations, with no initial correlations.

\begin{figure*}[h]
	\centering
	\includegraphics[width=0.97\textwidth]{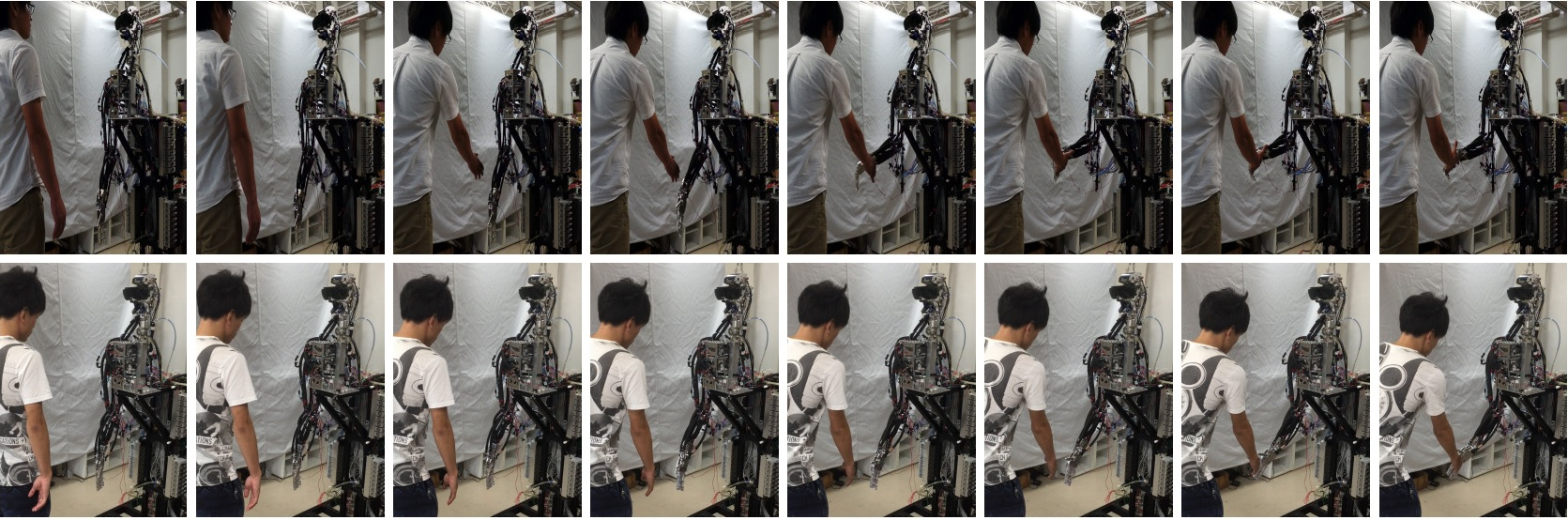}
	\caption{The motion of the human and robot over time when shaking hands at arbitrary end points with fast movement (top) and slow movement (bottom). Each sequence begins at the start of the interaction and each image is sampled at the same rate.}
	\label{fig:exp_handshake_images}
\end{figure*}

\begin{figure}
	\centering
    \includegraphics[width=0.7\linewidth]{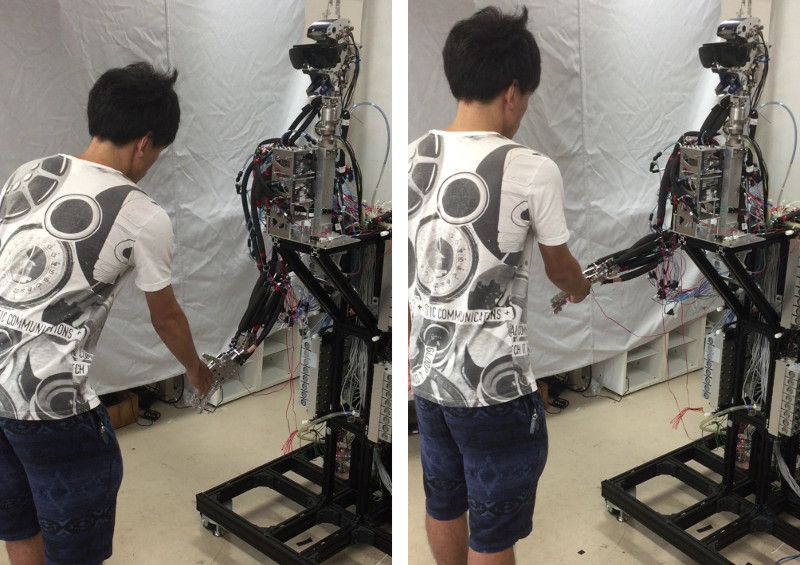}
	\caption{Different test interactions emphasizing BIP's spatial generalization. The left image shows a handshake low and closer to the robot, while the right image shows a handshake high and closer to the human.}
	\label{fig:handshake_space}
\end{figure}

\subsection{Musculoskeletal Robots}
\label{sec:challenges}

Robots which utilize pneumatic artificial muscles in a musculoskeletal configuration are desirable in the context of human-robot interaction.
Due to their air-driven nature, they are inherently back-drivable and thus well-suited for physical contact with humans.
Furthermore, when arranged in a musculoskeletal structure which mimics human anatomy they tend to produce predictable, legible~\cite{dragan2013legibility} motions.
This is important in interactions with humans as unexpected movements may result in injury or unsafe situations.

Despite the suitability of PAMs for human-robot interaction, there are various challenges in working with them.
The first relates to the nonlinear dynamics inherent to pneumatic actuation and compliant behavior. Once external forces are applied to the muscles, it becomes difficult to accurately determine from the actuation pressure where in space the actuated components are.
This limitation complicates control during interaction phases in which the robot experiences external forces, but it also introduces difficulties when training the robot. In learning from demonstration algorithms, a common technique to train the robot relies on kinesthetic guidance, \emph{i.e.} the robot is physically moved along a desired trajectory and the internal states of the actuators are recorded.
While it is possible to kinesthetically teach musculoskeletal robots actuated with PAMs~\cite{ikemoto2012direct}, this requires a specific design of the robot that is not always feasible. In the case of the robot used in this work, this approach was not possible since the state of pneumatic actuators is different when the robot is subject to external forces, \emph{i.e.} undergoing kinesthetic teaching, compared to when it is moving autonomously.

However, with BIP, only the correlations between the trajectories of the human and the robot need to be captured.
As such, during training, instead of adapting the robot's trajectory to the human's, we introduce a teaching method in which the human adapts to the robot while it is executing an open-loop policy.
More specifically, we created hand-crafted trajectories of the robot's desired action which are executed independently of the human during training, while the human produces an appropriate response.
In this way, we do not apply any external forces to the robot and are still able to accurately capture the relationship between the trajectories despite being unable to kinesthetically teach the robot.

Another consideration is that of the execution of the generated response trajectories.
The BIP algorithm updates at a given frequency and at each iteration generates a new response trajectory from the current point in the interaction (as estimated by $\phi$) to the end of the interaction; this is trivial to calculate with the estimated latent state representation.
This is preferable to generating only the next state, since depending on the size of the state dimension it may not be possible to calculate the next state in real-time.
However, this can produce discontinuous trajectories where the robot will need to make a large adjustment in position when a new response trajectory is generated; this is unsafe in human-robot interaction.
Therefore, an additional alpha-beta filter is employed to smooth the transition between the trajectories generated by BIP.

\begin{table}
	\centering
    \setlength\tabcolsep{5pt}
	\begin{tabular}{c  c  c  c  c  c  c}
		& \thead{Mean \\ (All)} & \thead{Var \\ (All)} & \thead{Mean \\ (T)} & \thead{Var \\ (T)} & \thead{Mean \\ (NT)} & \thead{Var \\ (NT)} \\
		\hline
        \hline
		\makecell{BIP \\ (Fast)} & \cellcolor{green!15} 0.2640 & 0.0013 & \cellcolor{gray!20} 0.2479 & 0.0001 & \cellcolor{green!15} 0.2752 & 0.0019 \\
		\hline
		\makecell{BIP \\ (Normal)} & 0.3094 & 0.0064 & \cellcolor{gray!20} 0.3148 & 0.0089 & \cellcolor{gray!20} 0.3062 & 0.0049 \\
		\hline
		\makecell{BIP \\ (Slow)} & 0.3824 & 0.0138 & 0.4049 & 0.0164 & 0.3689 & 0.0117 \\
		\hline
		\makecell{Static \\ (1)} & 0.3272 & 0.0055 & 0.2950 & 0.0012 & 0.3465 & 0.0070 \\
		\hline
		\makecell{Static \\ (2)} & 0.3009 & 0.0109 & \cellcolor{green!15} 0.2450 & 0.0018 & 0.3345 & 0.0134 \\
		\hline
		\makecell{Static \\ (3)} & 0.3387 & 0.0057 & 0.3430 & 0.0060 & \cellcolor{gray!20} 0.3365 & 0.0056 \\
		\hline
		\makecell{Static \\ (4)} & 0.2914 & 0.0017 & \cellcolor{gray!20} 0.2642 & 0.0030 & 0.3078 & 0.0002 \\
        \hline
        \hline
	\end{tabular}
	\caption{The mean Time-to-Completion (as a ratio of trajectory length, lower is better) and variance for all test participants (All), test participants who trained the model (T), and test participants who did not train the model (NT). BIP refers to our approach while static refers to an open-loop trajectory. Green cells indicate the scenarios with the smallest mean values while gray cells indicate scenarios that are not significantly different, as calculated with the Mann-Whitney U test with a $p$-value $< 0.05$.}
	\label{table:exp_ttc_results}
\end{table}

\begin{figure*}[th]
	\centering
	\includegraphics[width=0.97\textwidth]{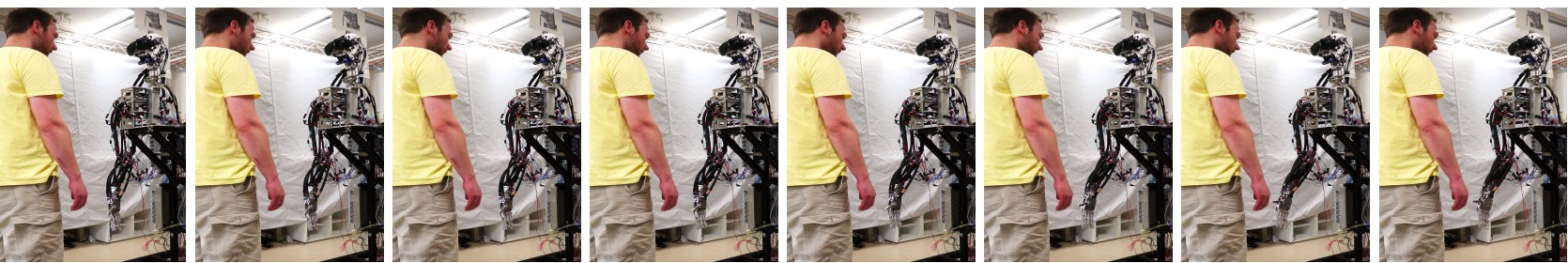}
	\caption{The motion of the human and robot over time when the human partner does not move their arm. The robot does not engage in a hand shake.}
	\label{fig:exp_handshake_images2}
\end{figure*}

\begin{figure*}
		\centering
		\includegraphics[width=0.99\linewidth, trim={4cm 0 4cm 0},clip]{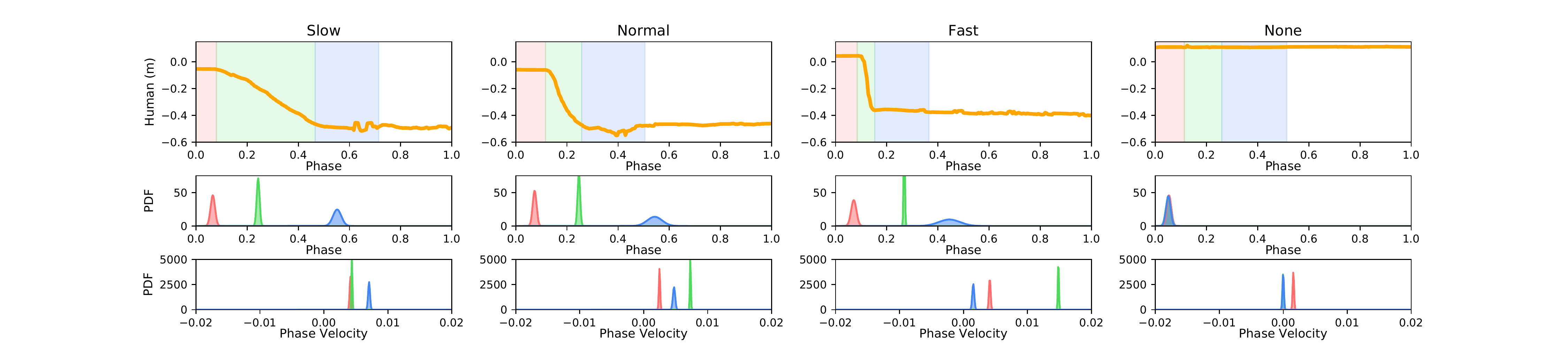}
	\caption{Top: the trajectory of the human's hand along the $x$-axis, which approximately corresponds to the distance to the robot, during different test interactions. The trajectory region shaded in red indicates the beginning portion of the interaction in which the participant has yet to move. The green region indicates the period in which the participant actively moves to shake the robot's hand. The blue region indicates the period after which the handshake is completed. Middle: the probability density corresponding to the estimated phase at the end of each aforementioned period (red, green, blue). Bottom: the probability density corresponding to the estimated phase velocity for each region.}
	\label{fig:distributions}
\end{figure*}

\section{Experiments and Results}
\label{sec:results}
In order to evaluate the effectiveness of our algorithm in interaction scenarios with a musculoskeletal robot, we designed an experiment in which a human performs a joint physical task with a robot. Specifically, we chose a handshake scenario and implemented a small set of manually-crafted trajectories for demonstrations. In this section we show that not only is BIP capable of reproducing a robust, legible handshake motion, but that it is capable of successfully generalizing to other humans.

\subsection{Experimental Setup}

The musculoskeletal robot~\cite{hitzmann2018anthropomorphic} employed in this work, shown in Fig.~\ref{fig:teaser}, contains $10$ kinematic degrees of freedom: $7$ in the arm linkage and $3$ in the shoulder mechanism.
These $10$ degrees of freedom are actuated with $27$ PAMs in an anatomical structure similar to that of humans.
Due to the prevalence of the spherical joints required to create a complex biomimetic structure, the robot does not contain conventional joint angle sensors. Rather, each PAM was equipped with a tension sensor and a pressure sensor to capture the state of the actuators, however, only the pressure sensors were used in this experiment. The PAMs themselves are connected to proportional valves which are controlled via PID controllers operating at $500$ Hz with a pressure reference signal. The values reported in this paper are measured in mPa as a difference from atmospheric pressure. Human subjects were tracked with $3$ degrees of freedom using skeleton tracking running on a Kinect v2 camera. The degrees of freedom correspond to the $x$-, $y$-, and $z$-position of the right hand, with the camera at an angle such that both the $x$- and $y$-axes indicate the distance and direction (left/right) of the handshake, while the $z$-axis indicates the height. Thus, the total number of degrees of freedom in this experiment was $30$ and sampling was performed at a rate of $30$ Hz.

During training, the robot executed $12$ manually-crafted trajectories with no feedback, i.e., in an open-loop fashion, as explained in Sec.~\ref{sec:challenges}. These trajectories were constructed via linear interpolation from a start pressure value and an end pressure value for each of the $27$ PAMs in the robot; for some of the PAMs, the start and end values were equal thus producing no movement for that DOF.
The end pressure values were chosen to produce handshake end points over the entire range of the robot in 3-D space.
The human participants who assisted in training were instructed to shake the hand of the robot once for each executed trajectory over a time window of $10$ seconds. Three participants contributed training demonstrations with $3$ repetitions of each trajectory, resulting in a total of $108$ total training demonstrations.
During testing, the three participants from training as well as five additional participants were asked to shake the hand of the robot in eight different scenarios.
In four of the scenarios, the robot once again executed a manually-crafted trajectory as in the training demonstrations, however, this time with different end points. In the remaining four scenarios, the BIP algorithm was employed with the robot generating response trajectories based on the human's hand movement.
The participants were asked to move their hand to an arbitrary location and shake the hand of the robot while moving their hand at a requested speed: fast, normal (as in the same speed as used in the demonstrations), slow, and a special case of no movement at all.
The speed definition was purposely vague and left up to the determination of each participant so as to adequately test the temporal robustness of the BIP algorithm.
Each test participant executed each scenario $2$ times, for a total of $128$ test trajectories.
A response trajectory was generated by BIP at a frequency of $3$ Hz (for computational reasons) using $15$ basis degrees for a total state dimension of $450$.
This trajectory was executed by the robot at $10$ Hz and consists solely of pressure values for each of the robot's degrees of freedom, no inverse kinematics or dynamics were used at any point as such models were unavailable.

\subsection{Results and Discussion}

\begin{figure}
		\centering
		\includegraphics[width=0.95\linewidth, trim={1cm 0 1cm 1cm},clip]{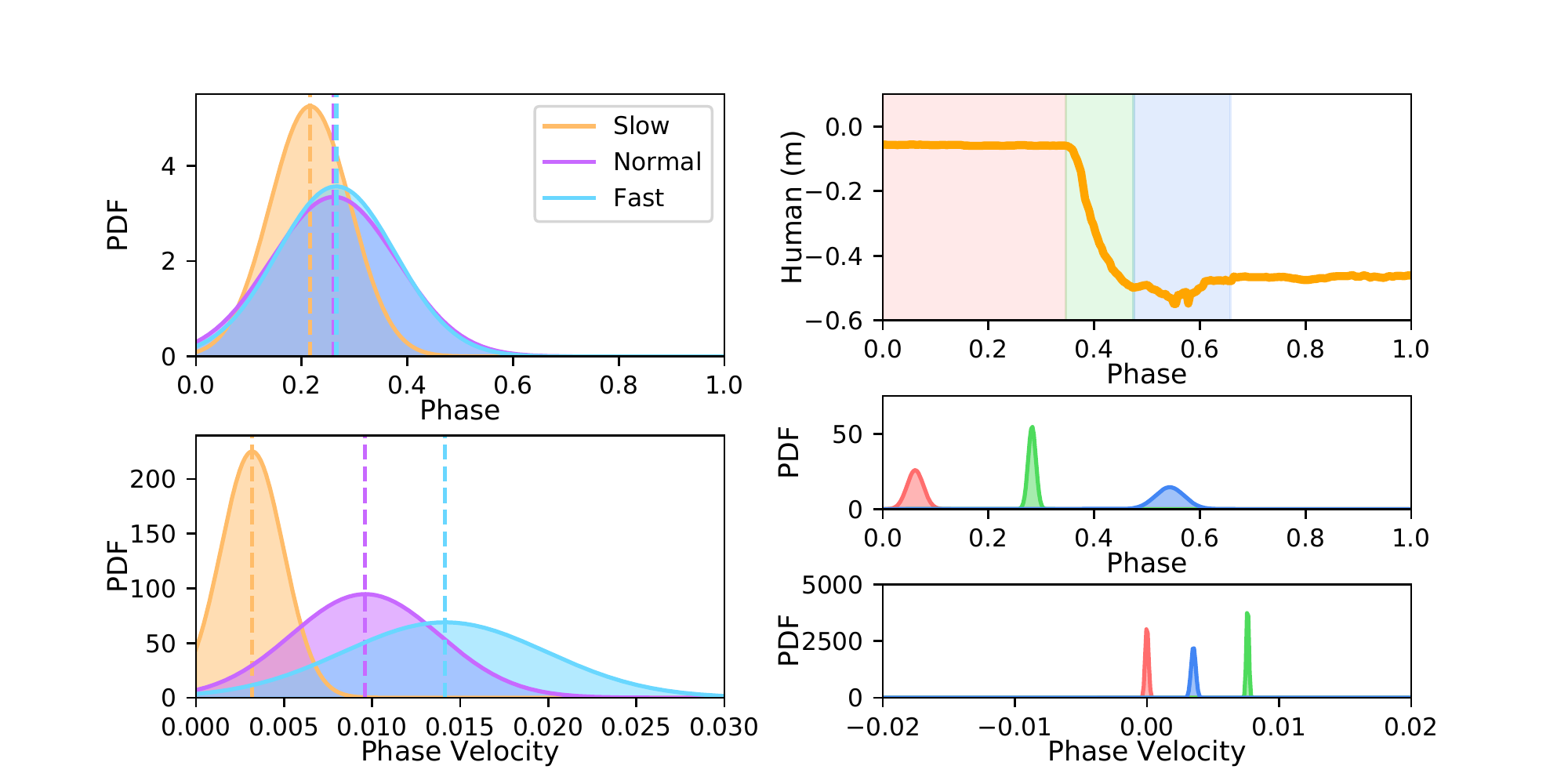}
	\caption{Left: the distribution of estimated phases (top) and phase velocities (bottom) after the participant moves to shake the robot's hand, for all tested slow, normal, and fast interactions. This corresponds to the green region shown in Fig.~\ref{fig:distributions}. Right: the same type of plot as in Fig.~\ref{fig:distributions} for the same normal speed interaction with the addition of an artificial pause at the beginning.}
	\label{fig:phase_distributions}
\end{figure}

\begin{figure}
		\centering
		\includegraphics[width=0.95\linewidth, trim={0.0cm 1.0cm 0.0cm 1.7cm},clip]{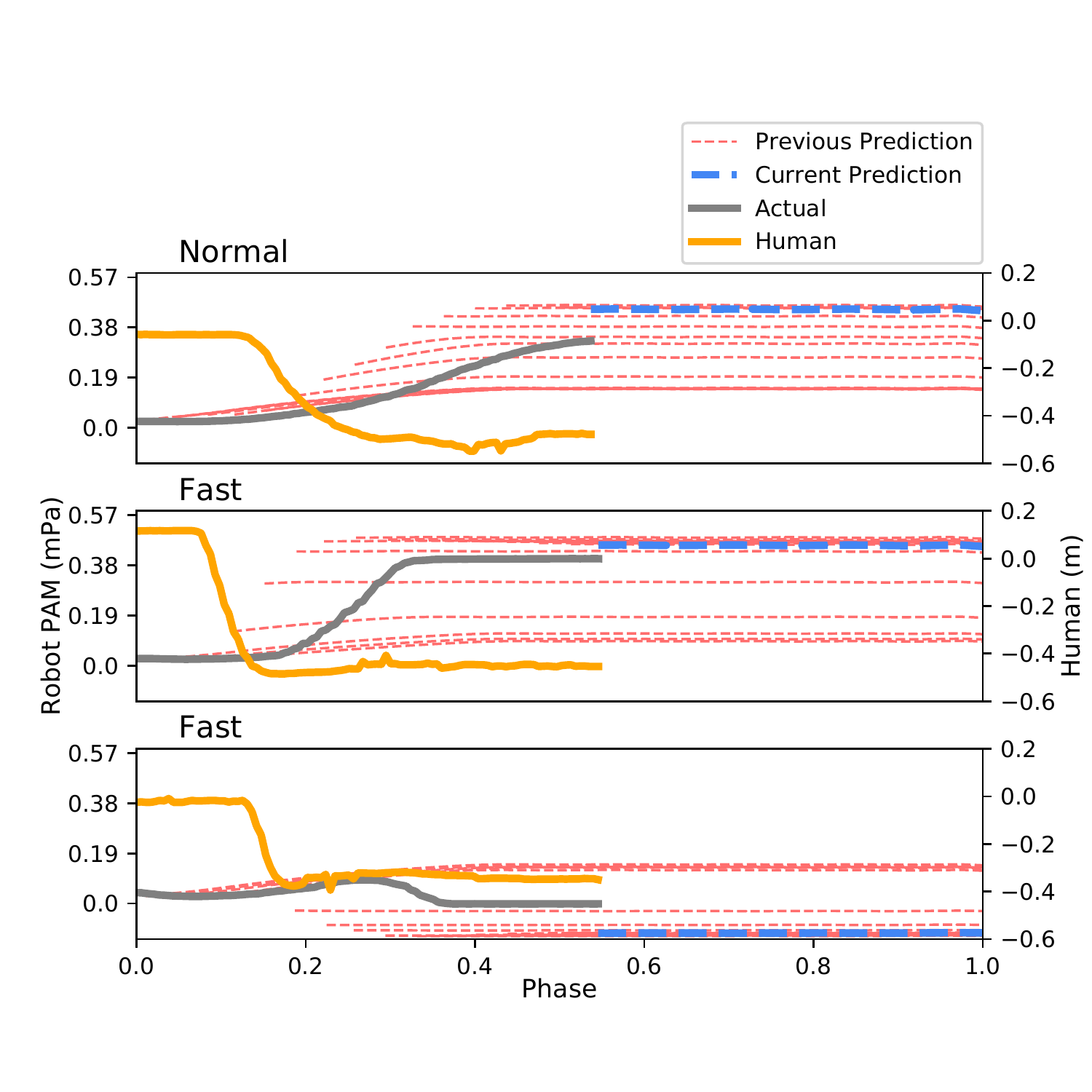}
	\caption{The predicted pressure trajectory for one of the robot's PAMs during normal (top) and fast (middle and bottom) test interactions. In all cases, the current prediction (blue) is from approximately 50\% through the interaction (the 17th inference step), with the predicted trajectories (red) from the previous 15 inference steps shown for reference. These trajectories are given to the robot's PID controller with the resulting actual pressure values shown in gray. The predicted and actual values do not directly coincide due to physical restrictions inherent to the pneumatic system that the learned model is unaware of. The corresponding human observations along the $x$-axis are shown in orange.}
	\label{fig:predicted_trajectory}
\end{figure}

\subsubsection{Qualitative Analysis}

A sequence of images showing two different test interactions over time for different participants is shown in Fig.~\ref{fig:exp_handshake_images}.
Qualitatively, these sequences simultaneously demonstrate three things: a) the robot learned to reproduce surprisingly human-like handshake motions despite the lack of access to any sort of analytical model, b) BIP is capable of generalizing this motion across both space \emph{and} time, and c) the algorithm is able to generalize across different human participants.
The temporal differences in particular can be clearly seen in Fig.~\ref{fig:exp_handshake_images}; in the fast (top) sequence, the human and robot have reached each other by the 5th frame, whereas in the slow (bottom) sequence, this does not occur until the 8th (last) frame.
While spatial differences are visible in these sequences, they can be more clearly seen in Fig.~\ref{fig:handshake_space}.
In the first image, the human chooses an end point for the handshake that is low and near the robot, while in the second image the same human participant chooses an end point that is higher and closer to the human.
Furthermore, an image sequence for an extreme edge case is shown in Fig.~\ref{fig:exp_handshake_images2} in which the human participant does not move their hand at all, thus, never beginning the interaction.
In response, the robot similarly does not proceed with the interaction and produces only minute movements.

\subsubsection{Temporal Analysis}

An analysis of the inferred phase at different points in the interaction for each movement speed is shown in Fig.~\ref{fig:distributions}.
The trajectories shown in the top plots represent the movement of the human's hand and are broken up into three periods: the period before the human has moved (red), the period during which the human is moving (green), and the period after the human has stopped moving (blue).
The phase and phase velocity predictions at the end of each period are shown such that the portion of the trajectory falling inside the corresponding shaded region has been observed.
These plots yield an interesting insight: the estimated phase for each movement speed is approximately the same as in the normal speed case.
That is, no matter how quickly the interaction proceeds in real time, the phase of the interaction is the same immediately before movement begins and immediately after.
Instead, it is the phase velocity that differs, particularly in the region during which movement occurs (the green region).
The slow speed interaction has the smallest phase velocity at the end of the movement period while the fast speed interaction has the greatest velocity.
The practical implication of this is that the amount of absolute time required to reach each point in the interaction differs due to different phase velocities, even if the phase value at each point is the same.
Since the state (Eq.~\ref{eq:ip_state_transition}) evolves according to a linear velocity model, a slower phase velocity requires more state updates to reach the same point in phase.

Figure~\ref{fig:phase_distributions} depicts the distribution of estimated phases and phase velocities for all interactions, and shows that this trend holds for all cases.
While the phase estimates exhibit little difference between the movement speed cases, the velocity estimates show a positive relationship between movement speed and phase velocity.
Analytically, this is due to our choice of initial uncertainty in the phase and phase velocity in Eq.~\ref{eq:bip_prior_cov}; the uncertainty in phase is set much lower than that of phase velocity because we are confident that the initial phase is $0$.
Phase and phase velocity have a non-zero covariance (an uncertainty in velocity affects the uncertainty in phase), but due to the different magnitudes in the initial uncertainty the phase velocity experiences more significant updates during inference.
The other observation we can make from Fig.~\ref{fig:phase_distributions} is that the larger phase velocity estimates exhibit greater variance.
This is due to the increased uncertainty corresponding to the larger estimates, which can also be seen in the green velocity plots of Fig.~\ref{fig:distributions}.
As a consequence, this impacts the uncertainty in the phase estimate for the blue region, as inference has already taken into account the velocities (and uncertainties) of the green region.
This is why the uncertainty for the blue region phase estimate increases with interaction speed in Fig.~\ref{fig:distributions}.

Lastly, we note that BIP handles the special case of no movement speed by reducing the phase velocity to $0$, thus stopping the temporal progression of the interaction.
This is visualized in the last column of Fig.~\ref{fig:distributions}.
However, this elimination of the temporal velocity is flexible and can be recovered from, as can be seen with the introduction of an artificial pause at the beginning of the interaction as shown in the right column of Fig.~\ref{fig:phase_distributions}.
Although the phase velocity has been reduced to $0$ at the end of the extended "no movement" (red) period, it quickly progresses when the human begins moving and yields phase estimates consistent with what we expect.
This indicates that not only is BIP robust to the movement speed of the interaction, but it is also robust to variations in \emph{when} the interaction begins.

\subsubsection{Spatial Analysis}

While we have demonstrated that the BIP framework is robust to temporal variations, our results also show that is also robust to spatial variations.
As Fig.~\ref{fig:predicted_trajectory} shows, the predicted robot trajectory is dependent both on the value of the human observations (spatial variation) as well as the rate of change (temporal variation).
The top and middle figures demonstrate spatial generalization through the pattern of increasing pressure values associated with the previous predictions (red dashed lines); as more human observations become available, BIP continues to refine the prediction.
However, because these two interactions have similar end points, the effect of movement speed becomes evident on the predictions.
While both predictions ultimately arrive at approximately the same pressure value for this particular PAM, BIP predicts these values for the fast interaction much earlier in response to the high rate of change of the human movement.
In contrast, the bottom figure is associated with a fast interaction at a different end point and the predicted trajectory is significantly different than the others, demonstrating the ability to spatially generalize to different end points.
These figures also highlight limitations of the BIP framework: namely, lack of knowledge of control lag and physical constraints.
The control lag can be observed via the phase shift between the predicted pressure values and when the actual robot actually reaches those values (gray line).
Similarly, BIP yields inferred pressure values that the system is incapable of reaching due to physical and/or mechanical limitations.
This is visible in the offset between the current prediction (blue dashed line) and the actual robot.

As a result of spatial generalization, empirically we hypothesize that the robot and human positions converge to the same physical location faster when both participants are active in the interaction (BIP scenarios), as opposed to only the human (static scenarios).
However, for technical reasons, we lack the position of the robot end-effector in 3D space and, therefore, measure the length of an interaction by how long it takes the human and robot degrees of freedom to converge to steady-state values, which we are referring to as Time-to-Completion.
To this end, the variance of each DoF was calculated over a sliding window of $2$ seconds.
If the variance of all DoFs within the window falls below a threshold -- $0.001$~m for the human and $0.001$ mPa for the robot -- then the start time of the sliding window is taken to be the Time-to-Completion.
These thresholds are chosen such that all scenarios yield a completion time.

\begin{figure}
	\centering
    \includegraphics[width=0.95\linewidth, trim={0cm 0cm 0cm 1.5cm},clip]{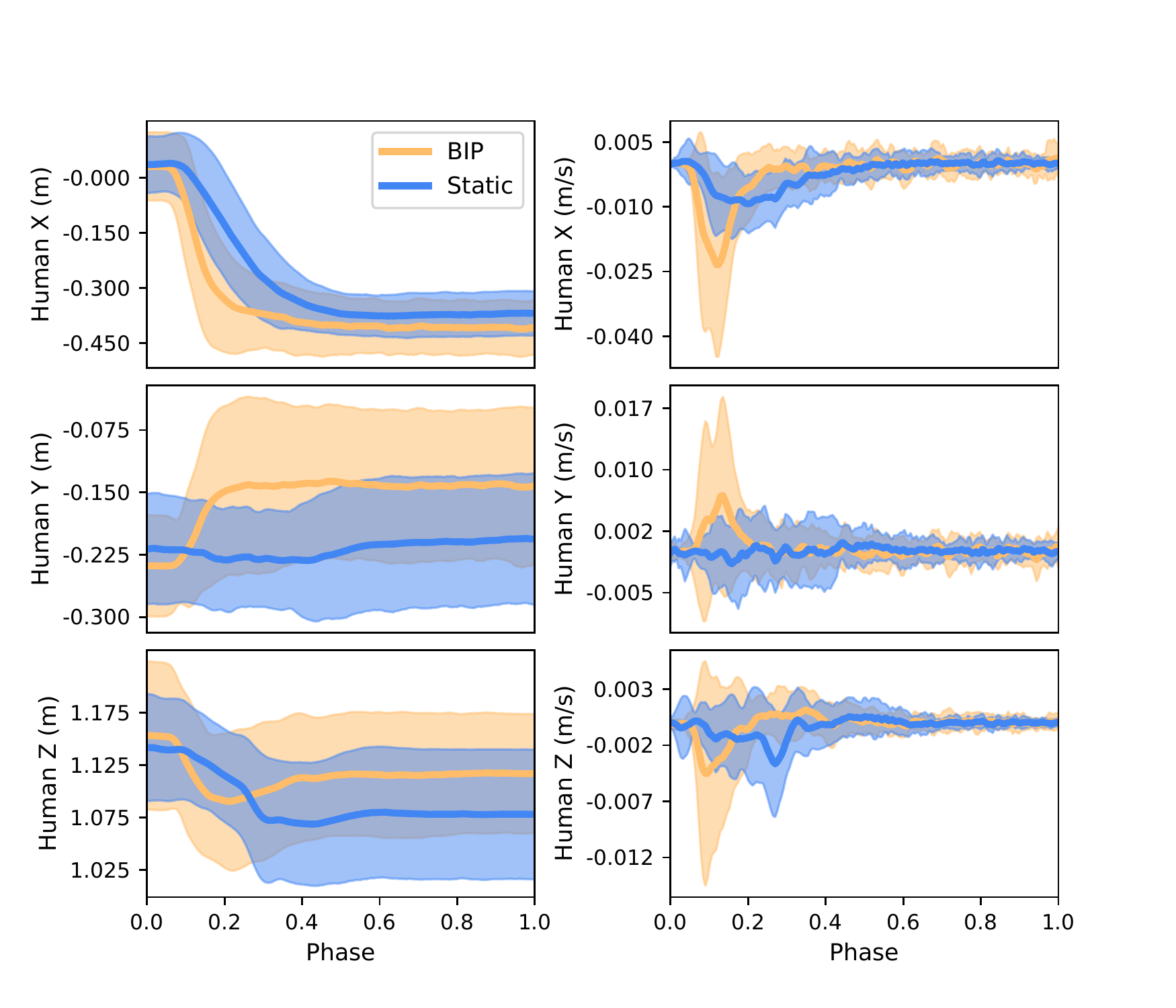}
	\caption{Left: the position distribution (mean and std dev) for all static and BIP scenarios for the human's hand along the $x$-axis (top), $y$-axis (middle), and $z$-axis (bottom). Right: the corresponding velocity distributions.}
	\label{fig:exp_velocity_dist}
\end{figure}

The results, shown in Table~\ref{table:exp_ttc_results}, support our hypothesis.
The mean Time-to-Completion values for all test participants are smaller when the BIP algorithm is employed compared to a static handshake trajectory.
The one exception is in the case of test participants who also trained the model (the T subset), in which case the BIP interactions were not statistically different than the second static trajectory.
This is an interesting observation, since it suggests that participants who had already participated in training the model were better able to predict where the robot would go when compared to new participants, despite new handshake end points.
Thus, their interactions resulted in lower Time-to-Completion times than the participants who hadn't trained the model (the NT subset).
However, despite their familiarity with the experiment, BIP was still able to produce similar completion values; in the case of new participants who hadn't trained the model, BIP resulted in more efficient interactions with smaller completion values.

The results in Table~\ref{table:exp_ttc_results} also indicate that the variance in Time-to-Completion is smaller for the static trajectories when compared to the BIP ones, particularly slow speed interactions.
This is expected considering that the test participants are free to choose their own handshake end point when testing with BIP as opposed to the static trajectories which result in an identical handshake end point for every participant.
This is especially clear when looking at the position and velocity distributions of the human as shown in Fig.~\ref{fig:exp_velocity_dist}.
Likewise, the human begins moving earlier in BIP scenarios than static ones, as they are dictating the end points rather than the robot -- visualized in the earlier peak of the velocity distributions.
The shape of the velocity distributions are also quite illuminating, as the BIP distributions closely resemble the typical bell-shaped velocity profiles of point-to-point movements observed in humans~\cite{flanagan1990trajectories}.

\begin{figure}
	\centering
	 \includegraphics[width=0.90\linewidth, trim={2.8cm 1cm 0.5cm 1.5cm},clip]{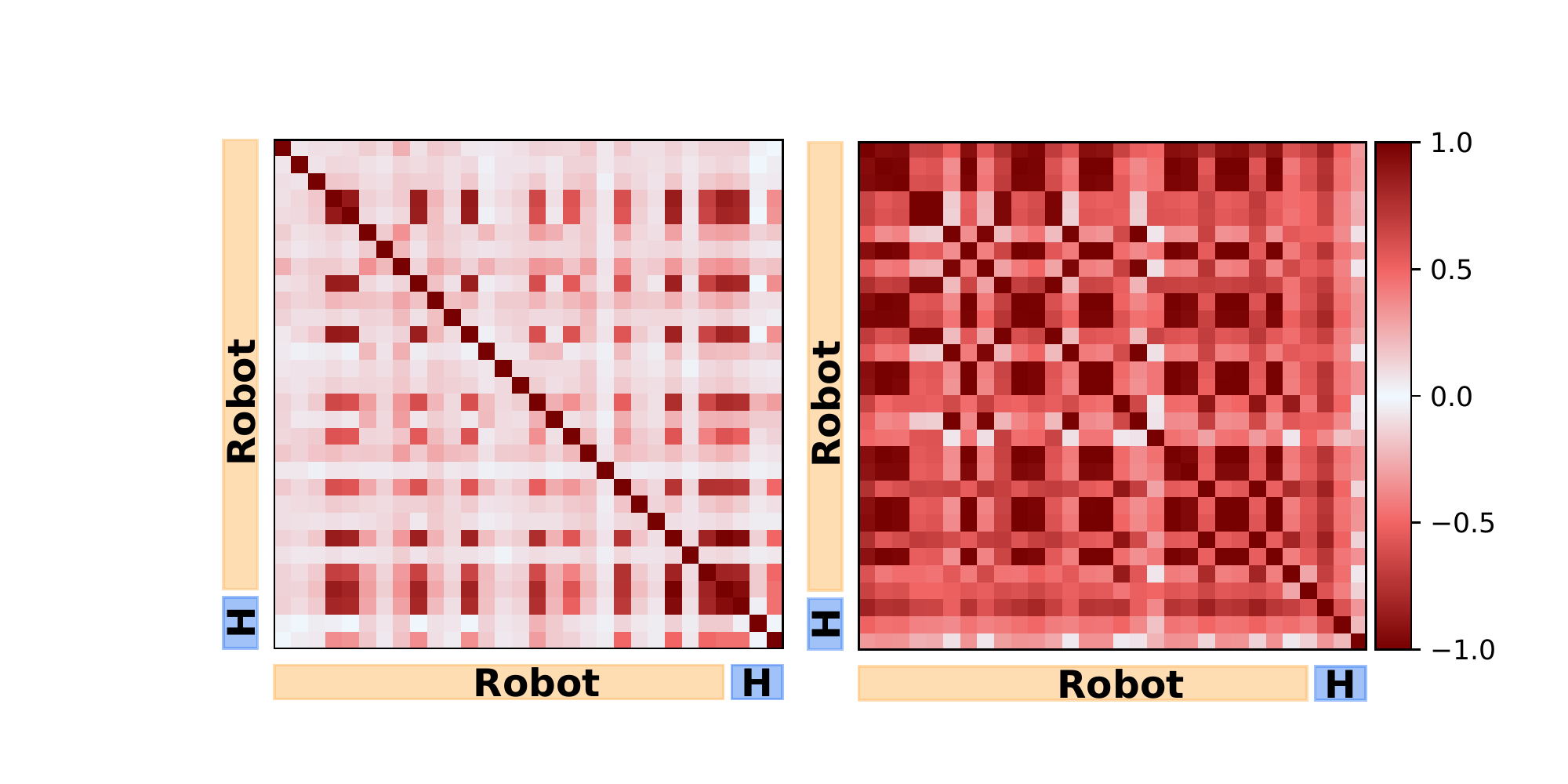}
	\caption{Mean Pearson correlation coefficients for static trajectories (left) and BIP (right). The color of the squares indicates the magnitude of the correlation. The first 27 rows and columns represent the coefficients for the robot degrees of freedom. The last 3 rows and columns represent the human. The mean correlations are calculated for all participants in all static scenarios and all BIP scenarios.}
	\label{fig:exp_corr}
\end{figure}

\begin{figure}
	\centering
    \includegraphics[width=0.95\linewidth, trim={0.3cm 0.8cm 0.3cm 0cm},clip]{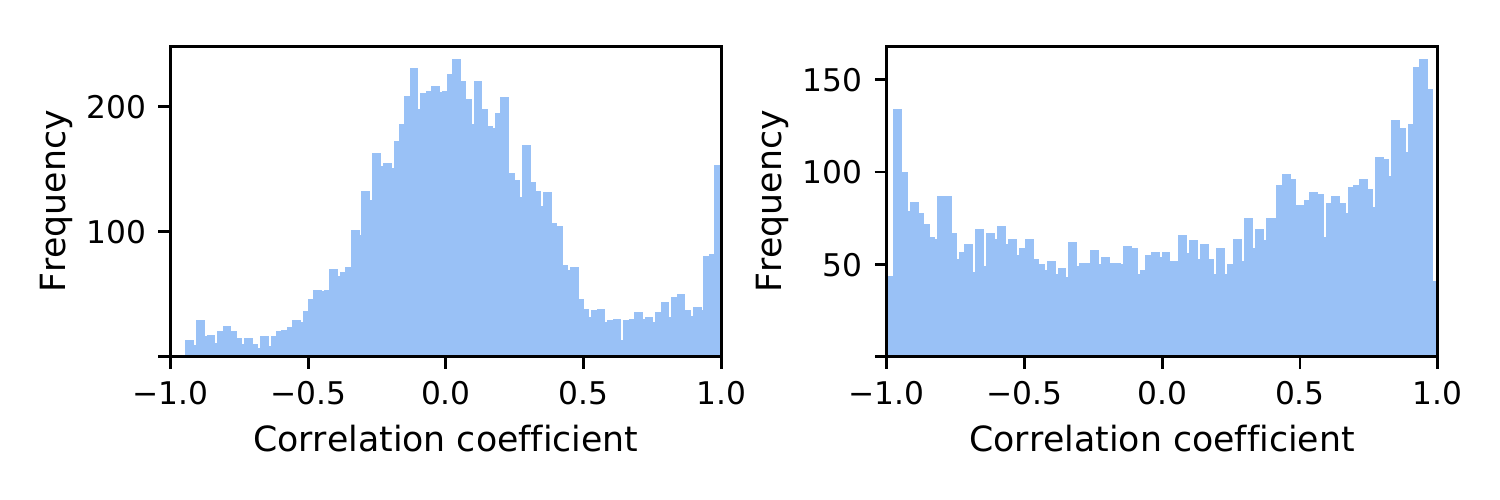}
	\caption{Histogram for the Pearson correlation coefficient generated from a sliding window over static trajectories (left) and BIP (right). This correlation is for an un-actuated PAM and hand position along the $y$-axis.}
	\label{fig:exp_histogram}
\end{figure}

The last result highlighting spatial generalization is that the Pearson correlation coefficients differ between the human and robot trajectories for BIP and static handshakes, as shown in Fig.~\ref{fig:exp_corr}.
Unsurprisingly, the only correlations above a magnitude of $0.5$ in the static scenarios are between the human degrees of freedom and the actuated robot degrees; the un-actuated degrees exhibit no significant correlations.
By contrast, all degrees of freedom which were actuated in the training process yielded significant correlations when BIP is used, indicating that BIP is effectively exploiting the model learned from demonstrations.
This is also supported by the histogram that is generated from a non-actuated degree of freedom (in the static scenarios) in Fig.~\ref{fig:exp_histogram}.
The histogram for the static scenarios is approximately Gaussian with a mean of $0$, in other words, Gaussian noise.
However, the histogram for the BIP scenarios is bi-modal with peaks at $-1$ and $1$, demonstrating strong positive and negative correlations.

\section{Conclusions}
\label{sec:conclusion}

In this work, we have shown that Bayesian Interaction Primitives can be successfully utilized in a physical, cooperative human-robot interaction scenario with a musculoskeletal robot. Despite the challenges inherent to pneumatic artificial muscles -- nonlinear dynamics and lack of kinesthetic teaching -- BIPs were able to learn strong spatiotemporal relationships between the movement trajectories of the robot and human test participants.
Furthermore, these relationships were generalizable to new human partners who did not take part in training the model, as well as significant temporal variations including an edge case in which the human does not interact at all.
At the same time, we have also identified limitations of the BIP framework, such as the disregard for control lag and mechanical constraints, which pose challenges to real-time, physical interactions.
In future work, we will study more complex interaction scenarios that involve a sequence of multiple primitives, introduce sensors of multiple modalities, and attempt to learn direct spatiotemporal relationships between human muscle activations and the PAMs of a musculoskeletal robot.


\section*{Acknowledgment}

We would like to thank Masuda Hiroaki for his assistance.
This work was supported by the National Science Foundation under Grant Nos.~1714060 and IIS-1749783, JSPS under KAKENHI Grant Number 18H01410, and the Honda Research Institute. J.C. is a JSPS International Research Fellow.


\bibliographystyle{IEEEtran} 
\bibliography{references}

\end{document}